\documentclass[11pt,letterpaper]{article}
\usepackage{naaclhlt2016}
\usepackage{times}
\pdfoutput=1

\usepackage{booktabs}
\usepackage{hyperref}
\usepackage{latexsym}
\usepackage{balance}
\PassOptionsToPackage{hyphens}{url}\usepackage{hyperref}

\naaclfinalcopy % Uncomment this line for the final submission
\def\naaclpaperid{***} %  Enter the naacl Paper ID here

\title{TwiSE at SemEval-2016 Task 4: Twitter Sentiment Classification}

\author{Georgios Balikas \and Massih-Reza Amini \\
        University of Grenoble-Alpes \\  {\tt firstnane.lastname@imag.fr}}
\date{}

\begin{document}

\maketitle

\begin{abstract}
This paper describes the participation of the team ``TwiSE'' in the SemEval 2016 challenge.
Specifically, we participated in Task 4, namely ``Sentiment Analysis in Twitter'' for which we implemented
sentiment classification systems for subtasks A, B, C and D.
Our approach consists of two steps. In the first step, we generate and validate diverse  feature sets for twitter sentiment evaluation, inspired by the work of participants of previous editions of such challenges. 
In the second step, we focus on the optimization of the evaluation measures of the different subtasks. To this end, we examine different learning strategies by validating them on the data provided by the task organisers. For our final submissions we used an ensemble learning approach (stacked generalization) for Subtask A and single linear models for the rest of the subtasks. In the official leaderboard we were ranked 9/35, 8/19, 1/11 and 2/14 for subtasks A, B, C and D respectively.\footnote{We make the code available for research  purposes at \url{https://github.com/balikasg/SemEval2016-Twitter\_Sentiment\_Evaluation}.}
% 
% The two remaining steps concern the learning part of our approach.
% In the second step, we investigate the performance of different learners (base learners) on the data provided by the challenge organisers.
% Then, in the final step, we propose to combine the outputs of the second step using an ensemble learning approach. The goal is to linearly combine the probabilistic outputs  of the base learners in order to generate our final predictions.
\end{abstract}

\section{Introduction}
During the last decade, short-text communication forms, such as Twitter microblogging, have been widely adopted and have become ubiquitous. 
Using such forms of communication, users share a variety of information. However, information concerning one's sentiment on the world around her has attracted a lot of research interest \cite{SemEval:2016:task4,rosenthal2015semeval}.

Working with such short, informal text spans poses a number of different challenges to the Natural Language Processing (NLP) and Machine Learning (ML) communities. Those challenges arise from the vocabulary used (slang, abbreviations, emojis) \cite{maas2011multi},  the short size, and the complex linguistic phenomena such as sarcasm \cite{rajadesingan2015sarcasm} that often occur. 

We present, here, our participation in Task 4 of SemEval 2016 \cite{SemEval:2016:task4}, namely Sentiment Analysis in Twitter. Task 4 comprised five different subtasks: Subtask A: Message Polarity Classification, Subtask B: Tweet classification according to a two-point scale, Subtask C: Tweet classification according to a five-point scale, Subtask D: Tweet quantification according to a two-point scale, and Subtask E: Tweet quantification according to a five-point scale. We participated in the first four subtasks under the team name ``TwiSE'' (\underline{Twi}tter \underline{S}entiment \underline{E}valuation). Our work consists of two steps: the preprocessing and feature extraction step, where we implemented and tested different feature sets proposed by participants of the previous editions of SemEval challenges \cite{tang-EtAl:2014:P14-1,kiritchenko2014nrc}, and the learning step, where we investigated and optimized the performance of different learning strategies for the SemEval subtasks. For Subtask A we submitted the output of a stacked generalization \cite{wolpert1992stacked} ensemble learning approach using the probabilistic outputs of a set of linear models as base models, whereas for the rest of the subtasks we submitted the outputs of single models, such as Support Vector Machines and Logistic Regression.\footnote{To enable replicability we make the code we used available at \url{https://github.com/balikasg/SemEval2016-Twitter\_Sentiment\_Evaluation}.}

% The objectives of our participation were twofold. First, since we were new to the sentiment analysis task, we tried to identify the main research directions and implement some of the proposed methods that deal with the problem. Second, we planned to research the deployment of an ensemble learning approach, that we only managed to fully developed for the first subtask. Our participation yielded:

The remainder of the paper is organised as follows: in Section 2 we describe the feature extraction and the feature transformations we used, in Section 3 we present the learning strategies we employed, in Section 4 we present a part of the in-house validation we performed to assess the models' performance, and finally, we conclude in Section 5 with remarks on our future work.

\section{Feature Engineering}
We present the details of the feature extraction and transformation mechanisms we used. Our approach is based on the traditional $N$-gram extraction and on the use of sentiment lexicons describing the sentiment polarity of unigrams and/or bigrams. 
For the data pre-processing, cleaning and tokenization\footnote{We adapted the tokenizer provided at \href{http://sentiment.christopherpotts.net/tokenizing.html}{http://sentiment.christopherpotts.net/tokenizing.html}} as well as for most of the learning steps, we used Python's Scikit-Learn \cite{scikit-learn} and NLTK \cite{BirdKleinLoper09}.

\subsection{Feature Extraction}
Similar to \cite{kiritchenko2014sentiment} we extracted features based on the lexical content of each tweet and we also used sentiment-specific lexicons. The features extracted for each tweet include:
\begin{itemize}
 \item N-grams with $N\in[1,4]$, character grams of dimension $M\in[3,5]$,
 \item \# of exclamation marks, \# of question marks, \# of both exclamation and question marks,
 \item \# of capitalized words and \# of elongated words, 
 \item \# of negated contexts; negation also affected the $N$-grams features by transforming a word $w$ in a negated context to $w\_ NEG$,
 \item \# of positive emoticons, \# of negative emoticons and a binary feature indicating if emoticons exist in a given tweet, and
 \item Part-of-speech (POS) tags \cite{gimpel2011part} and their occurrences partitioned regarding the positive and negative contexts.
 \end{itemize}
 
With regard to the sentiment lexicons, we  used:
\begin{itemize}
 \item manual sentiment lexicons: the Bing liu's lexicon \cite{hu2004mining}, the NRC emotion lexicon \cite{mohammad2010emotions}, and the  MPQA lexicon \cite{wilson2005recognizing},
 \item \# of words in positive and negative context belonging to the word clusters provided by the CMU Twitter NLP tool\footnote{\href{http://www.cs.cmu.edu/~ark/TweetNLP/}{http://www.cs.cmu.edu/~ark/TweetNLP/}}
  \item  positional sentiment lexicons: sentiment 140 lexicon \cite{go2009twitter} and
 the Hashtag Sentiment Lexicon \cite{kiritchenko2014sentiment}
 \end{itemize}
We make, here, more explixit the way we used the sentiment lexicons, using the Bing Liu's lexicon as an example. We treated the rest of the lexicons similarly. For each tweet, using the Bing Liu's lexicon we obtain a 104-dimensional vector. After tokenizing the tweet, we count how many words (i) in positive/negative contenxts belong to the positive/negative lexicons (4 features) and we repeat the process for the hashtags (4 features). To this point we have 8 features. We generate those 8 features for the lowercase words and the uppercase words.  Finally, for each of the 24 POS tags the \cite{gimpel2011part} tagger generates, we count how many words in positive/negative contenxts belong to the positive/negative lexicon. As a results, this generates $2\times8+24\times4=104$ features in total for each tweet.    

For each tweet we also used the distributed representations provided by \cite{tang-EtAl:2014:P14-1} using the min, max and average composition functions on the vector representations of the words of each tweet.

\subsection{Feature Representation and Transformation}
We describe the different representations of the extracted $N$-grams and character-grams we compared when optimizing our performance on each of the classification subtasks we participated. In the rest of this subsection we refer to both N-grams and character-grams as words, in the general sense of letter strings. We evaluated two ways of representing such features: (i) a bag-of-words representation, that is for each tweet a sparse vector of dimension $|V|$ is generated, where $|V|$ is the vocabulary size, and (ii) a hashing function, that is a fast and space-efficient way of vectorizing features, i.e. turning arbitrary features into indices in a vector \cite{weinberger2009feature}. We found that the performance using hashing representations was better. Hence, we opted for such representations and we tuned the size of the feature space for each subtask.

Concerning the transformation of the features of words, we compared the tf-idf weighing scheme and the $\alpha$-power transformation. The latter, transforms each vector $x=(x_1,x_2,\ldots,x_d)$ 
to $x'=(x_1^{\alpha},x_2^{\alpha},\ldots,x_d^{\alpha})$ \cite{Jegou}. The main intuition behind the $\alpha$-power transformation is that it reduces the effect of the most common
words. Note that this is also the rationale behind the $idf$ weighting scheme. However, we obtained better results using the $\alpha$-power transformation. Hence, we tuned $\alpha$ separately for each of the subtasks.

\section{The Learning Step}
Having the features extracted we experimented with several families of classifiers such as linear models, maximum-margin models, nearest neighbours approaches and trees. We evaluated their performance using the data provided by the organisers, which were already split in training, validation and testing parts. Table \ref{table:data} shows information about the tweets we managed to download. From the early validation schemes, we found that the two most competitive models were Logistic Regression from the family of linear models, and Support Vector Machines (SVMs) from the family of maximum margin models. It is to be noted that this is in line with the previous research \cite{mohammad2013nrc,buchner2015webis}. 
\begin{table*}\centering
 \begin{tabular}{lcccc}
 \toprule
  & Train & Development & DevTest & Test\\ 
 \midrule
 Subtask A & 5,500 & 1,831 & 1,791 & 32,009 \\
 Subtask B \& D & 4,346 & 1,325 & 1,417& 10,551\\
 Subtask C \& E & 5,482 &1,810 & 1,778 &20,632 \\
 \bottomrule
 \end{tabular}
\caption{Size of the data used for training and development purposes. We only relied on the SemEval 2016 datasets.}\label{table:data}
\end{table*}

\subsection{Subtask A}

Subtask A concerns a multiclass classification problem, where the general polarity of tweets has to be classified in one among three classes: ``Positive'', ``Negative'' and ``Neutral'', each denoting the tweet's overall polarity. The evaluation measure used for the subtask is the Macro-F$_1$ measure, calculated only for the Positive and Negative classes \cite{SemEval:2016:task4}.

\begin{table*}\centering
 \begin{tabular}{cccc}
 \toprule
 Algorithm & Multiclass approach & Optimizer & Probabilistic Outputs \\ 
 \midrule
 SVMs & crammer-singer & Liblinear & isotonic regression \\
 SVMs & crammer-singer & Liblinear & Platt's \\
 Logistic Regression & one-vs-all & Liblinear &  native \\
 Logistic Regression & multinomial loss function & LBFGS &  native \\
 \bottomrule
 \end{tabular}
\caption{Description of the base learners used in the stacked generalization approach.}\label{table:baseModels}
\end{table*}

Inspired by  the wining system of SemEval 2015 Task 10 \cite{buchner2015webis} we decided to employ an ensemble learning approach. Hence, our goal is twofold: (i) to generate a set of models that perform well as individual models, and (ii) to select a subset of models of (i) that generate diverse outputs and to combine them using an ensemble learning step that would result in lower generalization error.

We trained four such models as base models. Their details are presented in Table \ref{table:baseModels}. In the stacked generalization approach we employed, we found that by training the second level classifier on the probabilistic outputs, instead of the predictions of the base models, yields better results. Logistic Regression generates probabilities as its outputs. In the case of SVMs, we transformed the confidence scores into probabilities using two methods, after adapting them to the multiclass setting: the Platt's method \cite{platt1999probabilistic} and the isotonic regression \cite{zadrozny2002transforming}. To solve the optimization problems of SVMs we used the Liblinear solvers \cite{fan2008liblinear}. For Logistic Regression we used either Liblinear or LBFGS, with the latter being a limited memory quasi Newton method for general unconstrained optimization
problems \cite{yu2011dual}. To attack the multiclass problem, we selected among the traditional one-vs-rest approach, the crammer-singer approach for SVMs \cite{crammer2002algorithmic}, or the multinomial approach for Logistic Regression (also known as MaxEnt classifier), where the multinomial loss is minimised across the entire probability distribution \cite{malouf2002comparison}.

For each of the four base models the tweets are represented by the complete feature set described in Section 2. For transforming the n-grams and character-grams, the value of $\alpha\in\{0.2, 0.4, 0.6, 0.8, 1 \}$, the dimension of the space where the hashing function projects them, as well as the value of $\lambda\in\{ 10^{-7}, \ldots , 10^{6} \}$ that controls the importance of the regularization term in the SVM and Logistic regression optimization problems were selected by grid searching using 5-fold cross-validation on the training part of the provided data. We tuned each model independently before integrating it in the stacked generalization. 

Having the fine-tuned probability estimates for each of the instances of the test sets and for each of the base learners, we trained a second layer classifier using those fine-grained outputs. For this, we used SVMs, using the crammer-singer approach for the multi-class problem, which yielded the best performance in our validation schemes. Also, since the classification problem is unbalanced in the sense that the three classes are not equally represented in the training data, we assigned weights to make the problem balanced. Those weights were inversely proportional to the class frequencies in the input data for each class.    

\subsection{Subtask B}
Subtask B is a binary classification problem where given a tweet known to be about a given topic,
one has to classify whether the tweet conveys a positive or a negative sentiment towards the topic. The evaluation measure proposed by the organisers for this subtask is macro-averaged recall (MaR) over the positive and negative class. 

Our participation is based on a single model. We used SVMs with a linear kernel and the Liblinear optimizer. We used the full feature set described in Section 2, after excluding the distributed embeddings because in our local validation experiments we found that they actually hurt the performance. Similarly to subtask A and due to the unbalanced nature of the problem, we use weights for the  classes of the problem. Note that we do not consider the topic of the tweet and we classify the tweet's overall polarity. Hence, we ignore the case where the tweet consists of more than one parts, each expressing different polarities about different parts.

\subsection{Subtask C}
Subtask C concerns an ordinal classification problem. In the framework of this subtask, given a tweet known to be about a given topic, one has to estimate the sentiment conveyed by the tweet towards the topic on a five-point scale. Ordinal classification differs from standard multiclass classification in that the classes are ordered and the error takes into account this ordering so that not all mistakes weigh equally. In the tweet classification problem for instance, a classifier that would assign the class ``1'' to an instance belonging to class ``2'' will be less penalized compared to a classifier that will assign ``-1'' as the class . To this direction, the evaluation measure proposed by the organisers is the macroaveraged mean absolute error.  

Similarly to Subtask B, we submitted the results of a single model and we classified the tweets according to their overall polarity ignoring the given topics. Instead of using one of the ordinal classification methods proposed in the bibliography, we use a standard multiclass approach. For that we use a Logistic Regression that minimizes the multinomial loss across the classes. Again, we use weights to cope with the unbalanced nature of our data. The distributed embeddings are excluded by the feature sets.

We elaborate here on our choice to use a conventional multiclass classification approach instead of an ordinal one. We evaluated a selection of methods described in \cite{pedregosa2015feature} and in \cite{Gutierrez2015}. In both cases, the results achieved with the multiclass methods were marginally better and for simplicity we opted for the multiclass methods. We believe that this is due to the nature of the problem: the feature sets and especially the fine-grained sentiment lexicons manage to encode the sentiment direction efficiently. Hence, assigning a class of completely opposite sentiment can only happen due to a complex linguistic phenomenon such as sarcasm. In the latter case, both methods may fail equally. 

\subsection{Subtask D}
Subtask D is a binary quantification problem. In particular, given a set of tweets known to be about a given topic, one has to estimate the distribution of the tweets across the Positive and Negative classes. For instance, having 100 tweets about the new iPhone, one must estimate the fractions of the Positive and Negative tweets respectively. The organisers proposed a smoothed version of the Kullback-Leibler Divergence (KLD) as the subtask's evaluation measure. 

We apply a classify and count approach for this task \cite{bella2010quantification,forman2008quantifying}, that is we first classify each of the tweets and we then count the instances that belong to each class. To this end, we compare two approaches both trained on the features sets of Section 2 excluding the distributed representations: the standard multiclass SVM  and a structure learning SVM that directly optimizes KLD \cite{esuli2015optimizing}. Again, our final submission uses the standard SVM with weights to cope with the imbalance problem as the model to classify the tweets. That is because the method of \cite{gao2015tweet} although competitive was outperformed in most of our local validation schemes.

\section{The evaluation framework}

Before reporting the scores we obtained, we elaborate on our validation strategy and the steps we used when tuning our models. in each of the subtasks we only used the data that were realised for the 2016 edition of the challenge. Our validation had the following steps:
\begin{enumerate}                                                                                                                                                                                                                                \item Training using the released training  data,
\item validation on the validation data, 
\item validation again, in the union of the devtest and trial data (when applicable), after retraining on training and validation data.
\end{enumerate}
For each parameter, we selected its value by averaging the optimal parameters with respect to the output of the above-listed steps (2) and (3). It is to be noted, that we strictly relied on the data released as part of the 2016 edition of SemEval; we didn't use past data.

We now present the performance we achieved both in our local evaluation schemas and in the official results released by the challenge organisers. Table \ref{table:scores} presents the results we obtained in the ``DevTest'' part of the challenge dataset and the scores on the test data as they were released by the organisers. In the latter, we were ranked 9/35, 8/19, 1/11 and 2/14 for subtasks A, B, C and D respectively. Observe, that for subtasks A and B, using just the ``devtest'' part of the data as validation mechanism results in a quite accurate performance estimation. 

\begin{table}[t]\centering
 \begin{tabular}{lccr}
 \toprule
  &  DevTest & Test & Rank\\ 
 \midrule
 Subtask A & 56.01 & 58.61 & 9/35\\
 Subtask B & 0.748 & 0.748 & 8/18\\
 Subtask C & 0.8121 & 0.72  & 1/11\\
 Subtask D & 0.00018 & 0.053 & 2/14\\
 \bottomrule
 \end{tabular}
\caption{The performance obtained on the ``devtest'' data and the SemEval 2016 Task 4 test data.}\label{table:scores}
\end{table}

\section{Future Work}
That was our first contact with the task of sentiment analysis and we achieved satisfactory results. We relied on features proposed in the framework of previous SemEval challenges and we investigated the performance of different classification algorithms.

In our future work we will investigate  directions both in the feature engineering and in the algorithmic/learning part. 
Firstly, we aim to deal with tweets in a finer level of granularity. As discussed in Section 3, in each of the tasks we classified the overall polarity of the tweet, ignoring cases where the tweets were referring to two or more subjects. In the same line, we plan to improve our mechanism for handling negation. We have used a simple mechanism where a negative context is defined as the group of words after a negative word until a  punctuation symbol. However, our error analysis revealed that punctuation is rarely used in tweets. Finally, we plan to investigate ways to integrate more data in our approaches, since we only used this edition's data.

The application of an ensemble learning approach, is a promising direction towards the short text  sentiment evaluation. To this direction, we hope that  an extensive
error analysis process will help us identify better
classification systems that when integrated in 
our ensemble (of subtask A) will reduce the generalization error. 

\section*{Acknowledgments}
We would like to thank the organisers of the Task 4 of SemEval 2016, for providing the data, the guidelines and the infrastructure. We would also like to thank the anonymous reviewers for their insightful comments.

\balance
\bibliography{naaclhlt2016}

\bibliographystyle{naaclhlt2016}

\end{document}